\documentclass[journal]{IEEEtran}

\ifCLASSINFOpdf
\else
   \usepackage[dvips]{graphicx}
\fi
\usepackage{url}

\usepackage{graphicx}

\usepackage{caption}
\usepackage{subfigure}

\usepackage{bbding}
\usepackage{epsfig}
\usepackage{float}
\usepackage{graphicx}
\usepackage{amsmath,amsthm}               
{
	\theoremstyle{plain}
	\newtheorem{assumption}{Assumption}
}
\usepackage{amssymb}
\usepackage{amsthm}
\usepackage[linesnumbered,ruled]{algorithm2e}
\usepackage{colortbl,booktabs}
\usepackage{threeparttable}
\usepackage{subeqnarray}
\usepackage{cases}
\usepackage{multirow}
\usepackage{array}
\usepackage{bm}
\usepackage[switch]{lineno}
\usepackage{threeparttable}
\usepackage{fixltx2e}
\usepackage{enumerate}
\usepackage{epstopdf}

\newcolumntype{I}{!{\vrule width 2pt}}
\newlength\savedwidth

\newlength\savewidth

\newtheorem{definition}{Definition}

\theoremstyle{definition}

\usepackage{bm}

\begin{document}

\title{Preparation of Papers for IEEE Signal Processing Letters (5-page limit)}
\title{Visual Distribution Alignment and Uncertainty-Guided Belief Ensemble for Domain Generalization}
\title{Domain Generalization via Visual Alignment and Uncertainty-Guided Belief Ensemble}
\title{Mitigating Both Covariate and Conditional Shift for Domain Generalization}

\author{Jianxin Lin, Yongqiang Tang, Junping Wang and Wensheng Zhang

\IEEEcompsocitemizethanks{
	\IEEEcompsocthanksitem This work was supported by  the Key-Area Research and Development Program of Guangdong Province 2019B010153002, and by the  National Natural Science Foundation of China (62106266, U1936206). Yongqiang Tang and Wensheng Zhang are the corresponding authors.
	\IEEEcompsocthanksitem J. Lin, J. Wang and W. Zhang are with the Research Center of Precision Sensing and Control, Institute of Automation, Chinese Academy of Sciences, Beijing, 100190, China, and also with the School of Artificial Intelligence, University of Chinese Academy of Sciences, Beijing, 100049, China (e-mail: linjianxin2020@ia.ac.cn; junping.wang@ia.ac.cn; zhangwenshengia@hotmail.com).
	\IEEEcompsocthanksitem Y. Tang is with the Research Center of Precision Sensing and Control, Institute of Automation, Chinese Academy of Sciences, Beijing, 100190, China
	(e-mail: yongqiang.tang@ia.ac.cn).
}
}

\markboth{IEEE xxx, 2022}
{Shell \MakeLowercase{\textit{et al.}}: Bare Demo of IEEEtran.cls for IEEE Journals}
\maketitle

\begin{abstract}
Domain generalization (DG) aims to learn a model on several source domains, hoping that the model can generalize well to unseen target domains. The distribution shift between domains contains the covariate shift and conditional shift, both of which the model must be able to handle for better generalizability. In this paper, a novel DG method is proposed to deal with the distribution shift via Visual Alignment and Uncertainty-guided belief Ensemble (VAUE). Specifically, for the covariate shift, a visual alignment module is designed to align the distribution of image style to a common empirical Gaussian distribution so that the covariate shift can be eliminated in the visual space. For the conditional shift, we adopt an uncertainty-guided belief ensemble strategy based on the subjective logic and Dempster-Shafer theory. The conditional distribution given a test sample is estimated by the dynamic combination of that of source domains.
Comprehensive experiments are conducted to demonstrate the superior performance of the proposed method on four widely used datasets, {\it \textbf{i.e.}}, Office-Home, VLCS, TerraIncognita, and PACS.
\end{abstract}

% 100--175 words


\IEEEpeerreviewmaketitle

\section{Introduction}

\IEEEPARstart{C}{omputer} vision has made great progress in recent years with the help of deep learning under the basic assumption that all data are independently and identically distributed. However, in practical applications, images collected by different devices in different environments often follow different distributions. In such an out-of-distribution scenario, existing deep learning models suffer from the distribution shift and fail to generalize well  \cite{UDA-Long-2016}.

\begin{figure}[t]
	\setlength{\abovecaptionskip}{0pt}
	\setlength{\belowcaptionskip}{0pt}
	\renewcommand{\figurename}{Figure}
	\centering
	\includegraphics[width=0.4\textwidth]{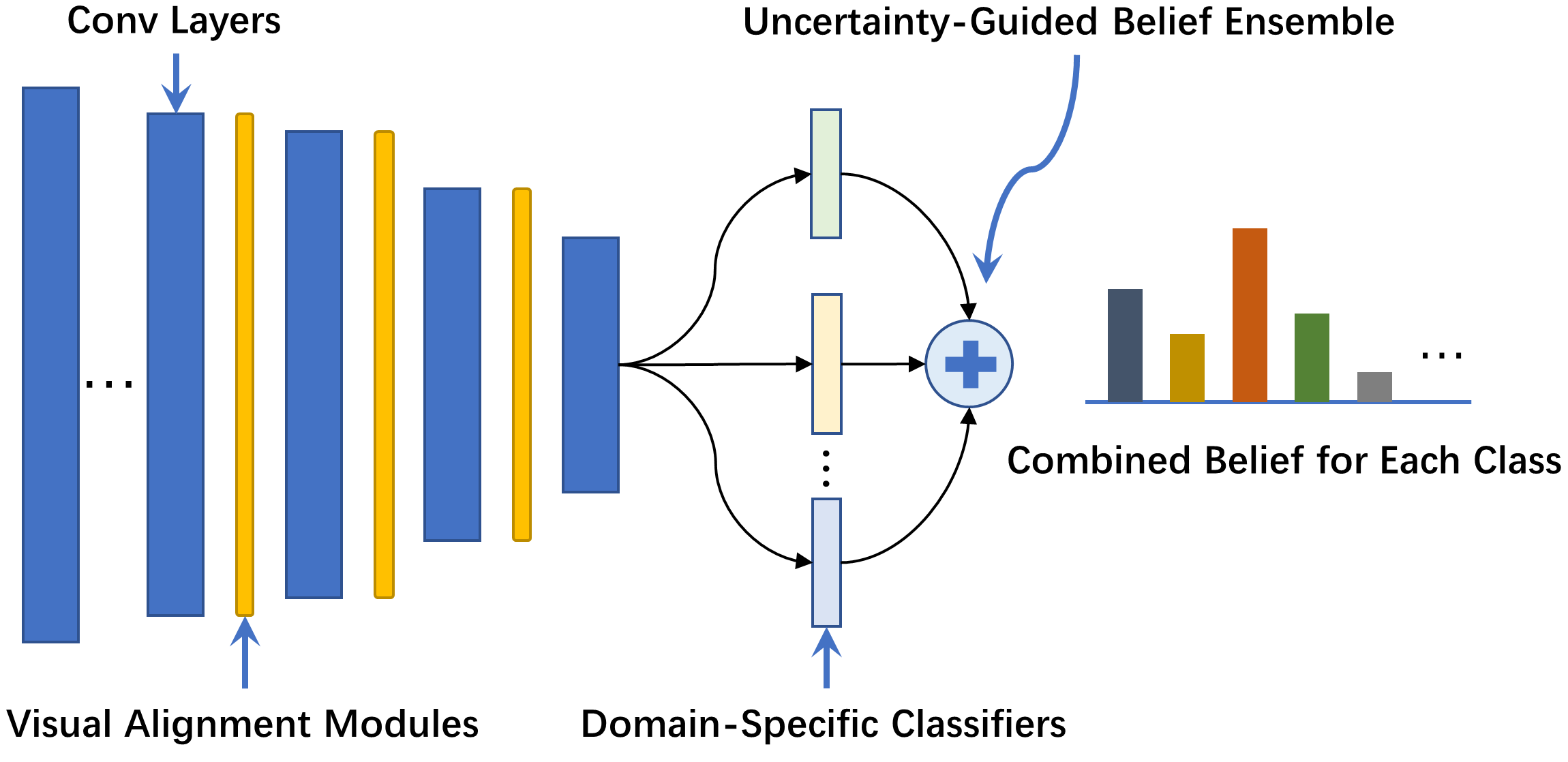}
	\caption{An overview of the proposed VAUE. A visual alignment module is proposed to align $P(X)$ in visual space via normalizing the style distributions. Uncertainty-guided belief ensemble is introduced to approximate $P(Y|X)$ with the combination of that of source domains at test time.}
	\label{fig}
\end{figure}

To tackle the distribution shift problem, great efforts have been made in Domain Adaptation (DA), which generally aims to transfer knowledge from a labeled source domain to an unlabeled target domain so that the learned model can perform well on the target domain \cite{UDA-Long-2016}. However, DA requires the target domain to be accessible which is hard to meet when the target domain changes dynamically. Furthermore, DA needs to retrain the model when applying it to another target domain, and it is time-consuming.
In recent years, Domain Generalization (DG) has attracted much attention, which tries to enable the model to generalize to unseen domains utilizing multiple source domains. Without access to target domains, DG improves the generalizability of models in out-of-distribution scenarios and has broad application prospects.

Among existing works, domain-invariant representation learning is a classic approach to DG. Let $X$ denote the input variable, {\it i.e.}, an image, and $Y$ denote the output variable, {\it i.e.}, a predicted label. As analysed in previous works, traditional models often suffer from covariate shift \cite{cs-pr-2012}, {\it i.e.}, $P^i(X) \ne P^j(X)$, and conditional shift \cite{Li_TDC-cian-eccv-2018}, {\it i.e.}, $P^i(Y|X) \ne P^j(Y|X)$ in out of distribution scenarios.
Some works \cite{ifr-wang-2013, bound-2021-arxiv} try to learn a representation space $Z = F(X)$ where the marginal distribution $P(Z)$ keeps the same across source domains so that the covariate shift can be eliminated assuming that $P(Y|X)$ keeps stable. Another line of research \cite{cdann-eccv-2018} tries to align the class-conditional distribution $P(Z|Y)$ in the representation space for a fine-grained distribution matching assuming that $P(Y)$ keeps stable. These works often hold impractical assumptions and fail to eliminate both the covariate and conditional shift. Additionally, though distributions of source domains are aligned, there is no guarantee that the distributions of unseen domains would be also aligned with that of source domains.

In this paper, we propose a new approach to eliminate both the covariate shift and conditional shift, as shown in Figure \ref{fig}. For the covariate shift, we propose to align $P(X)$ in visual space. In most scenarios, the domain shift is mainly caused by the image style which can be represented as feature statistics \cite{in-iccv-2017}. We attempt to model the common real-world style distribution, {\it i.e.}, the real-world distribution of feature statistics, which has yielded domain-specific style distributions with different selection biases. After that, we normalize all the domain-specific style distributions to the common style distribution so that the covariate shift is eliminated via visual alignment.
For the conditional shift, instead of aligning $P(Y|X)$ across source domains, we design a nonlinear ensemble scheme based on uncertainty modeling to dynamically approximate $P(Y | X=\mathbf{x}^t)$ given a test sample $\mathbf{x}^t$. The subject logic and Dempster-Shafer theory of evidence are first introduced to solve DG in our method. Comprehensive experiments have been conducted on four widely used datasets to demonstrate the effectiveness of our method.

\section{Proposed Method}

\subsection{Notations}
Let input sample and output label spaces be denoted as $\mathcal{X}$ and $\mathcal{Y}$ respectively. A domain is a set of data sampled from a joint distribution, which can be denoted as $\mathcal{D} = \{(\mathbf{x}_i, y_i)\}_{i=1}^n \sim P(X, Y)$, where $\mathbf{x} \in \mathcal{X} \subset \mathbb{R}^d, y\in \mathcal{Y} \subset \mathbb{R}$, and $P(X,Y)$ denotes the joint distribution of the sample and label. $X$ and $Y$ are the corresponding random variables. $C = |\mathcal{Y}|$ denoting the number of classes. Given $N$ source domains $\{ \mathcal{D}^i\}_{i=1}^N$ which follow different distributions, DG aims to learn a model which can generalize well on unseen target domains with unknown distribution shifts. In this paper, vectors are shown in bold, and the subscript indicates the corresponding dimension of the vector, {\it e.g.}, $\mathbf{v}$ and $v_i$.

% \subsection{Visual Alignment}
\subsection{Covariate Shift}

\begin{assumption}[\textbf{Independence Assumption}]
\label{ass:ia}
Let $f_{sem}(X) $ and $f_{sty}(X) $ be the semantic component and the style component extracted from $X$. $ P(f_{sem}(X))$ keeps stable across domains, while $P(f_{sty}(X))$ changes due to different selection biases on various domains.
$f_{sem}(X)$ is independent with  $f_{sty}(X)$ so that $P(f_{sem}(X), f_{sty}(X))  = P(f_{sem}(X))\times P(f_{sty}(X)) $.

\end{assumption}

Under Assumption \ref{ass:ia}, we can align $\{P^i(X)\}_{i=1}^N$ of source domains by normalizing $\{P^i(f_{sty}(X))\}_{i=1}^N$ to the same target distribution.
According to the previous work \cite{in-iccv-2017}, we know that feature statistics of the features in intermediate layers of deep networks, {\it i.e.}, means and standard deviations computed on each feature channel, have encoded the style information of images. 
Given an image, the intermediate feature after a certain layer of a network is denoted as $\mathbf{z} \in \mathbb{R}^{K\times H\times W}$.
%, where $B$ denotes the batch size, $C$ denotes the 
The vector of feature statistics $\bm{\mu} \in \mathbb{R}^{K}$ and  $\bm{\sigma} \in \mathbb{R}^{K}$ can be calculated as follows:

\begin{equation}
	\begin{aligned}
		{\mu}_{k}(\mathbf{z}) & = \frac 1 {HW} \sum_{h=1}^H \sum_{w=1}^W{z}_{khw}\\
		\sigma_{k}(\mathbf{z}) & = \sqrt{\frac 1 {HW}\sum_{h=1}^H \sum_{w=1}^W ({z}_{khw} - \mu_{k}(\mathbf{z}))^2}
	\end{aligned}
\end{equation}
As analysed in Assumption \ref{ass:ia}, we can approximately represent $f_{sty}(X)$ with $\{\bm{\mu}, \bm{\sigma}\}$. Hence we can normalize the distribution of $\{\bm{\mu}, \bm{\sigma}\}$ of images to align $P(X)$ in visual space. Assuming that all domain-specific feature statistics are sampled from the real-world style distribution with various selection biases, given a batch of images equally sampled from each domain, we aggregate all samples to approximate the real-world style distribution with an empirical distribution of feature statistics $P^{emp}(f_{sty}(X))$.
$P^{emp}(f_{sty}(X))$ is formalized as a Gaussian distribution. Its parameters are calculated as follows:
\begin{equation}
	\begin{aligned}
		\mathbf{m}_\mu & = \frac 1 B \sum_{b=1}^{B} \bm{\mu}(\mathbf{z}^{b}), \ \mathbf{m}_\sigma = \frac 1 N \sum_{b=1}^{B} \bm{\sigma}(\mathbf{z}^b)\\
		\mathbf{\Sigma}_\mu & = \frac 1 {B-1} \sum_{b=1}^B(\bm{\mu}(\mathbf{z}^b) - \mathbf{m}_\mu)(\bm{\mu}(\mathbf{z}^b) - \mathbf{m}_\mu)^T\\
		\mathbf{\Sigma}_\sigma & = \frac 1 {B-1} \sum_{b=1}^N(\bm{\sigma}(\mathbf{z}^b) - \mathbf{m}_\sigma)(\bm{\sigma}(\mathbf{z}^b) - \mathbf{m}_\sigma)^T
	\end{aligned}
\end{equation}
where $B$ denotes the batch size. Then we can get that:
\begin{small} 
\begin{equation}
	P^{emp}\left(f_{sty}(X) = \{\bm{\mu}_s, \bm{\sigma}_s\}\right) = \mathcal{N}( \bm{\mu}_s|\mathbf{m}_\mu, \mathbf{\Sigma}_\mu) \times \mathcal{N}( \bm{\sigma}_s| \mathbf{m}_\sigma, \mathbf{\Sigma}_\sigma) 
\end{equation}
\end{small}

We normalize $\{P^i(f_{sty}(X))\}_{i=1}^N$ to $P^{emp}(f_{sty}(X))$ for the marginal distribution alignment. For each image, we replace the original feature statistics with new values, $\bm{\sigma}_{s}$ and $\bm{\mu}_{s}$, sampled from the empirical distribution:
\begin{gather}
\{{\bm{\mu}}_{s}, \bm{\sigma}_{s}\} \sim P^{emp}(f_{sty}(X)) \\
\mathbf{z}  = ((\mathbf{z} - \bm{\mu}(\mathbf{z})) / \bm{\sigma}(\mathbf{z}) ) \odot \bm{\sigma}_{s} + \bm{\mu}_{s}
\end{gather}
where $/$ and $\odot$ denote channel-wise division and multiplication respectively.
By normalizing the distribution of feature statistics to the empirical distribution $P^{emp}(f_{sty}(X))$, which is a reasonable substitute for the real-world style distribution, the visual style distributions  $P(f_{sty}(X))$ of source domains are aligned. The operation of visual alignment is a plug-and-play module, which can be flexibly inserted into the neural networks at different positions like Batch Normalization layers. Each visual alignment module is randomly enabled according to a probability factor $\tau$ to control the ratio of samples to be transformed.

% \subsection{Uncertainty-Guided Belief Ensemble}
\subsection{Conditional Shift}

To overcome the conditional shift, we also design an uncertainty-guided belief ensemble strategy to dynamically approximate $P(Y|X)$ given a test sample based on the Subjective Logic (SL) and Dempster-Shafer theory of evidence. For more details, we refer readers to published works \cite{edl-nips-2018}, \cite{sl-2018}, \cite{ce-2002}, \cite{tmv-pami-2022}.
\subsubsection{Subjective Logic} 
We construct a linear domain-specific classifier for each source domain following a shared feature extractor, which is expected to fit the domain-specific posterior distribution. Instead of the softmax operation, we choose $\exp(\cdot)$ to make outputs of linear classifiers non-negative. The unnormalized non-negative outputs of each domain-specific classifier, $\{\mathbf{e}^i\}_{i=1}^N, \mathbf{e}^i \in \mathbb{R}^C$, can be regarded as the collected {\it evidences} in favor of a sample to be classified into a certain category.
Following SL, we define the belief masses, $\{\mathbf{b}^i\}_{i=1}^N$, $\mathbf{b}^i \in \mathbb{R}^{C}$, and uncertainty masses, $\{u^i\}_{i=1}^N$, $u^i \in \mathbb{R}$, for $i$-th domain-specific classifier as follows:
\begin{equation}
	\begin{aligned}
		\mathbf{b}^i = \frac {\mathbf{e}^i}{S^i},\ u^i = \frac {C} {S^i},\ \text{and}\ u^i + \sum_{c=1}^C b_c^i = 1
	\end{aligned}
\end{equation}
where $S^i = \sum_{c=1}^C(e^i_c + 1)$. Under this definition, the uncertainty mass is inversely proportional to the total evidence. 

The output of classical deep learning classifiers is a probability assignment over all classes. SL parameterizes a Dirichlet distribution with the evidence, which is a probability density function for all possible probability assignments over all classes. In other words, Dirichlet distribution is defined on a $C$-dimensional unit simplex, $\mathcal{S}_C = \{\mathbf{p} | \sum_{c=1}^C p_c = 1\ \text{and}\ \forall c,\  p_c \ge 0\}$ , where every point is a $C$-dimensional probability assignments. Specifically, the parameters of Dirichlet distribution for $i$-th domain-specific classifier is defined as $\bm{\alpha}^i =\mathbf{e}^i +1 $, and then the $i$-th Dirichlet distribution can be denoted as:
\begin{equation}
	\text{Dir}(\mathbf{p} \mid \boldsymbol{\alpha}^i)= \begin{cases}\frac{1}{B(\boldsymbol{\alpha}^i)} \prod_{c=1}^{C} p_{c}^{\alpha^i_{c}-1} & \text { for } \mathbf{p} \in \mathcal{S}_{C}, \\ 0 & \text { otherwise ,}\end{cases}
\end{equation}
where $B({\cdot})$ is a $C$-dimensional multinomial beta function. 

Under above definitions, given a sample, all domain-specific classifiers will output their evidence collected from the sample,  $\{\mathbf{e}^i\}_{i=1}^N$. And then belief masses $\{\mathbf{b}^i\}_{i=1}^N$ for each category and uncertainty masses $\{u^i\}_{i=1}^N$  can be derived. What is more, Dirichlet distributions for each domain-specific classifier $\{\text{Dir}(\mathbf{p}| \bm{\alpha}^i)\}_{i=1}^N$ will be formalized with derived evidences. 
%By this way, the model can provide a more reliable measure of uncertainty than vanilla softmax operation.

\subsubsection{Reduced Dempster's Combinational Rule}
We adopt a reduced Dempster's combinational rule \cite{tmv-pami-2022} to nonlinearly combine the predictions of all domain-specific classifiers.

\begin{definition}[\textbf{Reduced Dempster's Combinational Rule}]
	\label{def:rdc} 
	Given two sets of masses $\mathcal{M}^1=\{\mathbf{b}^1, u^1\}$ and $\mathcal{M}^2=\{\mathbf{b}^2, u^2\}$, the combination $\mathcal{M}=\{\mathbf{b}, u\}$ can be calculated as follows:
	\begin{equation}
		b_c = \frac 1 {1-F}(b^1_c b^2_c + b^1_c u^2 + b^2_c u^1), u = \frac 1 {1-F} u^1 u^2,
	\end{equation}
	where $F = \sum_{i\ne j}b_i^1b_j^2$ reflecting the conflict between two mass sets. The combination can be denoted as $\mathcal{M} = \mathcal{M}^1\oplus \mathcal{M}^2$.
\end{definition}
All mass sets given by domain-specific classifiers can be combined as $\mathcal{M} = \mathcal{M}^1\oplus \mathcal{M}^2\oplus \cdots \oplus \mathcal{M}^N$. By doing so, we can formalize the overall Dirichlet distribution based on the combined belief masses and uncertainty mass $\{\mathbf{b}, u\}$. Specifically, parameters of the combined Dirichlet distribution can be derived as follows: $
	S = \frac C u, \mathbf{e} = \mathbf{b} \times S\ \text{and}\ \bm{\alpha} = \mathbf{e} + 1$.

\subsubsection{Single-Domain and Cross-Domain Training}
\label{cross domain}
After illustrating the definitions of {\it evidence}, {\it belief}, {\it uncertainty} and {\it combination rule}, we now specifically show the detailed training process.
Given an input sample, the $i$-th domain-specific classifiers will output a mass set $\mathcal{M}^i$ and a Dirichlet distribution $\text{Dir}(\mathbf{p} | \bm{\alpha}^i) $. 
Let $\mathbf{x}$ denote an input sample, and $\mathbf{y}$ denote the corresponding one-hot label. To train this classifier with $\mathbf{x}$, the loss function is designed as follows \cite{edl-nips-2018}:
\begin{equation}
	\begin{aligned}
		&\mathcal{L}_{ece}(\mathbf{x}, \mathbf{y},\bm{\alpha}^i) \\
		= & \mathbb{E}_{\text{Dir}(\mathbf{p} | \bm{\alpha}^i)}\left[\sum_{c=1}^{C}-y_{c} \log \left(p_{c}\right)\right]+\lambda \text{KL}[\text{Dir}(\mathbf{p} | \bm{\tilde\alpha}^i) || \text{Dir}(\mathbf{p} | \mathbf{1})] \\
		= & \sum_{c = 1}^C y_c \left(\psi(S^i) - \psi(\alpha^i_c)     \right) +\lambda \text{KL}[\text{Dir}(\mathbf{p} | \bm{\tilde\alpha}^i) || \text{Dir}(\mathbf{p} |\mathbf{1} )] ,
	\end{aligned}
\end{equation}

where $\bm{\tilde\alpha}^i = \mathbf{y} + (\mathbf{1}-\mathbf{y}) \odot \bm{\alpha}^i$, $\mathbf{1}$ is a vector with all elements  equal to 1,  and $\psi(\cdot)$ is the digamma function. In this paper, $\lambda$ is set to 0.01. The first term is an expectation of cross entropy computed over the Dirichlet distribution which is essentially a Bayes risk. And the second term is proposed to enforce the evidence for incorrect labels to shrink to 0 \cite{edl-nips-2018}.

To model the uncertainty well, we design a single-domain training part and a cross-domain training part.
For the single-domain part, the data of $i$-th domain are only fed into $i$-th domain-specific classifier and compute the loss $\mathcal{L}_{ece}$. For the cross-domain part, the data of $i$-th domain are fed into all domain-specific classifiers except the $i$-th domain. After that, $N-1$ mass sets are combined. And the combined Dirichlet distribution is derived, parameters of which are denoted as $\bm{\alpha}^{/i}$. Hence the loss function can be designed as follows:
\begin{equation}
	\mathcal{L}_{D}= \frac 1 N \sum_{i=1}^N \frac 1 {|\mathcal{D}^i|} \sum_{j=1}^{|\mathcal{D}^i|}\left(\mathcal{L}_{ece}(\mathbf{x}^j, \mathbf{y}^j, \bm{\alpha}^i) + \mathcal{L}_{ece}(\mathbf{x}^j, \mathbf{y}^j, \bm{\alpha}^{/i}) \right)
\end{equation}
%For a better effectiveness of the ensemble of classifiers, we enforce the correlation across different dimensions of the extracted feature before classifiers to shrink to 0. By doing so, we hope different domain-specific classifiers could focus more on different dimensions so that the diversity of classifiers could be enhanced. 

Furthermore, we enforce the correlation across different dimensions of the extracted feature to shrink to 0. We found that this design can prevent the numerical computation problem of the proposed method during training.
Specifically, for a batch of features $\mathbf{z} \in \mathbb{R}^{B \times d}$, the mean of features $m = \frac 1 B \sum_{b=1}^B \mathbf{z}^b$, then the decorrelation loss can be designed as:
\begin{equation}
\mathcal{L}_{decor} = \left|\frac 1 {B-1} \left(\mathbf{z} - m\right)^T\left(\mathbf{z} - m\right) \odot (\mathbf{1}-I) \right|_1
\end{equation}
% $\mathcal{L}_{decor} = \left|\frac 1 {B-1} \left(\mathbf{z} - m\right)^T\left(\mathbf{z} - m\right) \odot (1-I) \right|_1$, 
where $I$ is a identity matrix.
The finally loss function can be designed as :
\begin{equation}
	\mathcal{L} = \mathcal{L}_{D} + \mathcal{L}_{decor}
\end{equation}

%$\mathcal{L} = \mathcal{L}_{D} + \mathcal{L}_{decor} $.

\subsubsection{Testing}
At test time, a test image is fed into the feature extractor firstly, and then the extracted feature is fed into all domain-specific classifiers. After that, all mass sets produced by classifiers are combined based on the reduced Dempster's combinational rule. The class which has the highest combined belief mass is the final prediction. The combined uncertainty mass shows the overall confidence of the prediction.
Given a test sample, the real working labeling function is the nonlinear combination of that of source domains, which automatically adjusts the weight of each labeling function of source domains according to the uncertainty of the sample at test time. By this way the conditional shift between train data and test data is eliminated.

\section{Experimental Results}
\label{sec:guidelines}

\subsection{Experimental Setup}
\subsubsection{Datasets}
For demonstrating the effectiveness of the proposed method VAUE,  we evaluate it on four widely used DG datasets, namely 
{\it Office-Home} (4 domains, 65 classes, and 75,588 images),
{\it VLCS} (4 domains, 5 classes, and 10,729 images),
{\it TerraIncognita} (4 domains, 10 classes, and 24,788 images),
{\it PACS} (4 domains, 7 classes, and 9,991 images). 
For all experiments, one domain is selected as the unseen test domain, and the others are treated as training domains.

\subsubsection{Implementation Details}
For all experiments, the networks, which are pre-trained on ImageNet, are trained by AdamW with a batch size of 64 for each domain and weight decay of 5e-4. The batch normalization is frozen during the training. The exponential moving average of model parameters with a momentum of 0.999 is conducted to make the training processes more stable. We adopt the standard data augmentations following \cite{domainbed-iclr-2021}. The visual alignment modules are inserted after 1,2,3-th ConvBlock of ResNet. All results are reported based on the average top-1 classification accuracy over three repetitive runs.

For Office-Home, VLCS, and TerraIncognita, following \cite{domainbed-iclr-2021}, we randomly split each training domain into 8:2 training/validation splits. All validation splits of training domains are aggregated as an overall validation set, which is used for model selection. The probability factor $\tau$ of visual alignment modules is set to 0.1. For Office-Home and TerraIncognita, the models are trained with a learning rate of 1e-5 for up to 5k iterations. For VLCS, models are trained for up to 2k iterations. We summarize the results of comparison methods reported in \cite{domainbed-iclr-2021} in Table \ref{table:Office-Home50}.
For PACS, the original train-validation split provided by \cite{pacs-iccv-2017} is adopted for a fair comparison with more diverse and novel competitors. The probability factor $\tau$ is set to 1. The models are trained with a learning rate of 1e-4 for up to 4k iterations.
We summarize the results of competitors reported in original papers in Table \ref{table:PACS}.

\begin{table}[t]
	\renewcommand{\arraystretch}{1.}
	\centering
	{
		\caption{Performance comparison on Office-Home, VLCS, and TerraIncognita dataset with ResNet-50 as backbone.}
		\label{table:Office-Home50}
		\setlength{\tabcolsep}{1.mm}{\scalebox{0.69}{
			\begin{threeparttable}
				\begin{tabular}{l | c c c c| c | c c c c| c | c c c c| c }
					\toprule[1.5pt]
					\multicolumn{1}{l |}{Dataset} & \multicolumn{5}{c|}{Office-Home}  & \multicolumn{5}{c|}{VLCS} & \multicolumn{5}{c}{TerraIncognita} \\
					\midrule
					Domain  &   A  & C   &  P   &   R    &  Avg.  
					&   C  & L   &  S   &   V    &  Avg.  
					&   L100  & L38   &  L43   &   L46    &  Avg.  \\
					\midrule
					ERM \cite{domainbed-iclr-2021}  &$ 61.3$       & $52.4 $      & $75.8$     &$ 76.6   $ & $66.5 $                 
					& $97.7$        &$ 64.3$        & $73.4$       & $74.6 $      & $77.5$                 
					& $49.8$       &$ 42.1$       & $56.9$       & $35.7 $      & $46.1 $                \\
					IRM  \cite{irm-2019-arxiv}                & $58.9$     & $52.2 $     &$ 72.1$    & $74.0$       &$ 64.3$                 
					&$98.6 $     &$ 64.9$       & $73.4 $       & $77.3$       & $78.5 $                
					&$ 54.6 $       &$ 39.8$       &$ 56.2  $     &$ 39.6$        &$ 47.6 $                \\
					DRO \cite{dro-arxiv-2019}   &$ 60.4  $     & $52.7$     &$75.0 $      &$ 76.0 $      & $66.0$         
					& $97.3$       & $63.4 $       &$ 69.5$        &$ 76.7 $       & $76.7 $                
					&$ 41.2$       & $38.6 $      & $56.7  $      & $36.4$        & $43.2 $                \\
					Mixup \cite{mixup-iclr-2018}  & $62.4 $    & ${54.8} $     &$ {76.9} $      & $78.3$        & $68.1 $                
					&$ 98.3 $       &$ 64.8$        & $72.1$        & $74.3$        & $77.4 $                
					& $\underline{59.6} $       & $42.2$        &$ 55.9 $       & $33.9 $       & $47.9 $                \\
					MLDG  \cite{mldg-aaai-2018} &$ 61.5 $      & $53.2$      & $75.0$      & $77.5 $      & $66.8  $               
					& $97.4$        & $65.2$        & $71.0$        &$ 75.3$        & $77.2 $                
					& $54.2 $       &$ \underline{44.3} $       & $55.6 $       &$ 36.9$        & $47.7 $                \\
					CORAL \cite{coral-iccv-2019}   &$ {65.3}$     & $54.4  $     &$ 76.5$      & ${78.4}$      & ${68.7}  $               
					& $98.3$       &$ \textbf{66.1}$       &$ 73.4$        & $\underline{77.5} $       & $78.8 $                
					&$ 51.6 $       &$ 42.2$        &$ 57.0 $      & $39.8 $       & $47.6 $                \\
					MMD \cite{mmd-icml-2015}   & $60.4$        &$ 53.3 $      & $74.3$      & $77.4$        & $66.3 $                
					& $97.7$        & $64.0  $      &$ 72.8 $       &$ 75.3 $       & $77.5 $                
					&$ 41.9 $       &$34.8 $       & $57.0 $       & $35.2  $      & $42.2 $                \\
					DANN \cite{dann-icml-2015} & $59.9$      & $53.0$   &$73.6 $      & $76.9$      & $65.9$               
					& $\textbf{99.0}$        &$ 65.1 $       &$ 73.1 $       &$ 77.2 $       &$ 78.6$                 
					& $51.1$        & $40.6 $       & $57.4$        &$ 37.7 $       &$ 46.7 $                \\
					CDANN \cite{cdann-eccv-2018}  &$ 61.5$     &$ 50.4$     & $74.4$       &$ 76.6$       & $65.8$               
					& $97.1$        &$ 65.1$      &$ 70.7 $       & $77.1$       & $77.5$                 
					& $47.0$       & $41.3$        & $54.9$        & $39.8 $      &$ 45.8$                 \\
					MTL \cite{mtl-jmlr-2021} &$ 61.5$      &$ 52.4$    & $74.9$      & $76.8$       & $66.4$                
					& $97.8 $       & $64.3 $       &$ 71.5 $      &$75.3$       & $77.2$                 
					&$ 49.3 $       & $39.6 $       & $55.6 $      &$ 37.8 $       & $45.6$                 \\
					SagNet	\cite{sagnet-cvpr-2021}  & $63.4$   & ${54.8}$  & $75.8$   & $78.3$  & $68.1 $      
					&$ 97.9$        &$ 64.5 $       &$ 71.4 $       &$ \underline{77.5}$        &$ 77.8 $                
					&$ 53.0 $       & $43.0$        &$ 57.9 $       &$ 40.4$        & $48.6$                 \\
					ARM \cite{arm-nips-2021}  & $58.9$    &$ 51.0$      & $74.1$      & $75.2$      & $64.8$                 
					& $98.7$        &$ 63.6 $       & $71.3 $       &$ 76.7 $       & $77.6 $                
					& $49.3$        & $38.3 $       & $55.8 $       & $38.7$        & $45.5 $                \\
					VREx \cite{re-icml-2021} &$ 60.7$      &$ 53.0$      & $75.3$      & $76.6$       & $66.4$                
					& $98.4 $       &$ 64.4 $       &$ \underline{74.1}$        & $76.2 $      &$ 78.3   $              
					&$ 48.2$        &$ 41.7$        &$ 56.8$        & $38.7 $      & $46.4 $                \\
					RSC \cite{rsc-eccv-2020}			 &$ 60.7$     & $51.4$       &$ 74.8$      & $75.1$     &$ 65.5$                
					& $97.9 $       &$ 62.5  $     & $72.3  $      &$ 75.6 $       &$ 77.1  $               
					&$ 50.2 $       &$ 39.2$        &$ 56.3$        & $\underline{40.8} $       &$ 46.6 $              \\
					
					Fishr \cite{fishr-icml-2022} &$62.4$     & $54.4$       &$76.2$      & $78.3$     &$67.8$                
					&$\underline{98.9}$     & $64.0$       &$71.5$      & $76.8$     &$ 77.8$                  
					&$50.2$     & $43.9$       &$55.7$      & $39.8$     &$47.4$                \\
					
					ITL-NET \cite{ltl-icml-2022}  &$\underline{65.6}$     & $\underline{55.6}$       &$ \underline{77.5}$      & $\underline{78.6}$     &$ \underline{69.3}$                
					&$98.3$     & $\underline{65.4}$       &$ \textbf{75.1}$      & $76.8$     &$ \underline{78.9}$                  
					&$58.4$     & $\textbf{46.2}$       &$ \underline{58.5}$      & $\textbf{40.9}$     &$ \textbf{51.0}$                \\
					
					\midrule
					VAUE (ours)  & $\textbf{67.8}$   & $\textbf{59.9}$ & $\textbf{78.1}$ & $\textbf{79.8}$ & $\textbf{71.4}$ 
					& $\textbf{99.0}$   & $ 64.7$ & $\textbf{75.1} $ & $\textbf{79.4}$ & $\textbf{79.6}$ 
					& $\textbf{59.7}$   & $41.0$ & $\textbf{59.2}$ & ${39.0}$ & $\underline{49.7}$ \\
					\bottomrule[1.5pt]
				\end{tabular}
				\footnotesize
				The best result is in bold face. Underlined ones represent the second-best results.
	\end{threeparttable}}}}
\end{table}

\subsection{Results}
As shown in Table \ref{table:Office-Home50}, we evaluate the proposed method VAUE on three datasets with ResNet-50 as the feature extractor. We can see that VAUE achieves the best average accuracy on Office-Home and VLCS and the second average accuracy on TerraIncognita. The domain shift on TerraIncognita is less about the image style. So the performance improvement is not as obvious as that on other datasets.
We also evaluate VAUE on PACS dataset with ResNet-18 and ResNet-50 as backbone respectively to compare our method with a greater variety of methods, as shown in Table \ref{table:PACS}. We can see that VAUE achieves the best average accuracy with both ResNet-18 and ResNet-50.
We notice that there is an obvious performance drop on the Photo (P) domain compared to DeepAll, which minimizes the empirical risk by aggregating samples from all source domains. This is mainly due to the ImageNet pretraining \cite{randconv-iclr-2020}. The images of the Photo (P) domain are highly similar to those of ImageNet. If the training strategy is changed, this benefit of pre-training may be reduced.

\subsection{Ablation Study}
To better demonstrating the effectiveness of the proposed VAUE, we conduct an ablation study by constructing four variant methods as shown in Table \ref{table:ablation}. VAUE w/o VA is a variant of VAUE without visual alignment modules. VAUE w/o EC is a variant constructed by replacing the reduced Dempster's combinational rule with a vanilla average combination. VAUE w/o CD is a variant that trains the models without cross-domain training mentioned in Section \ref{cross domain}. VAUE w/o UE is a variant constructed by removing the whole part of the uncertainty-guided belief ensemble.
We can see that all four designs provide significant performance improvement to the final accuracy. We note that the visual alignment modules produce much performance gain except on Photo domain. Because after being aligned to a common Gaussian distribution, the resulting image style could be more different from that of ImageNet.

\begin{table}[t]
	\renewcommand{\arraystretch}{1.}
	\centering
	{
		\caption{Performance comparison on PACS dataset with ResNet-18 and ResNet-50 as backbone.}
		\label{table:PACS}
		\setlength{\tabcolsep}{1.4mm}{\scalebox{0.75}{
			\begin{threeparttable}
				\begin{tabular}{l | c c c c |c | c c c c | c }
					\toprule[1.5pt]
					\multicolumn{1}{ l |}{Backbone} & \multicolumn{5}{c |}{ResNet-18} & \multicolumn{5}{c }{ResNet-50} \\
					\midrule
					Domain  &   {A}  & { C}   &   { P}   &  { S}    &   { Avg.}    &   { A}  & { C}   &   { P}   &  { S}    &   { Avg.} \\
					\midrule
					DeepAll \cite{dger-nips-2020}  & $78.93$  & $75.02$ & $96.60$  & $70.48$ & $80.25$  
					& $86.18$  & $76.79$ & $98.14$  & $74.66$ & $83.94$  \\	
					DSON \cite{dson-eccv-2020}   & ${84.67}$         & $77.65$          & $95.87$          & $\underline{82.23}$          & $85.11$          
					& $87.04$         & $80.62$          & $95.99$          & $82.90$          & $86.64$   \\
					DMG \cite{dmg-eccv-2020}  & $76.90$          & $\underline{80.38}$          & $93.35$          & $75.21$          & $81.46$         
					& $82.57$          & $78.11$          & $94.49$          & $78.32$          & $83.37$		 \\
					EISNet  \cite{eisnet-eccv-2020} & $81.89$         & $76.44$          & $95.93$          & $74.33$          & $82.15$  
					& $86.64$         & $81.53$          & $97.11$          & $78.07$          & $85.84$  \\
					DGER \cite{dger-nips-2020}  \ \    & $80.70$         & $76.40$          & ${96.65}$          & $71.77$     & $81.38$    
					& ${87.51}$         & $79.31$          & $\underline{98.25}$          & $76.30$          & $85.34$         \\
					RSC \cite{rsc-eccv-2020}			& $83.43$         & $80.31$          & $95.99$          & $80.85$          & $85.15$    
					& ${87.89}$         & $82.16$          & $97.92$          & $83.35$          & ${87.83}$    \\
					
					RSC* \cite{rsc-eccv-2020}			& $78.90$         & $76.88$          & $94.10$          & $76.81$          & $81.67$    
					& $81.38$         & $80.14$          & $93.72$          & $82.31$          & $84.38$    \\
					
					MDGHybrid \cite{mdghybrid-icml-2021} & $81.71$         & $\textbf{81.61}$      & $\underline{96.67}$      & $81.05$          & $\underline{85.53}$   
					& $86.74$         & ${82.32}$      & $\textbf{98.36}$      & $82.66$          & $87.52$    \\
					MGFA \cite{mgfa-bmvc-2021}	  & $81.70$         & $77.61$          & $95.40$          & $76.02$          & $82.68$          
					& $86.40$         & $79.45$          & $97.86$          & $78.72$          & $85.60$          \\
					FACT	\cite{fact-cvpr-2021}	 & $\underline{85.37}$         & $78.38$          & $95.15$          & $95.15$          & $84.51$    
					& $\underline{89.63}$         & $81.77$          & $96.75$          & $\underline{84.46}$          & $\underline{88.15}$    \\
					pAdaIN \cite{padin-cvpr-2021}  & $81.74$         & $76.91$          & $96.29$          & $75.13$          & $82.51$    
					& $85.82$         & $81.06$          & $97.17$          & $77.37$          & $85.36$    \\		
					EFDMix \cite{efd-cvpr-2022}    & $83.90$         & $79.40$          & $\textbf{96.80}$          & $75.00$          & $83.90$    
					& $\textbf{90.60}$         & ${82.50}$          & $98.10$          & $76.40$          & $86.90$    \\		
					DSFG  \cite{dsfg-wacv-2022} & $83.89$         & $76.45$      & $95.09$      & $78.26$          & $83.42$    
					& $87.30$         & $80.93$      & $96.59$      & ${83.43}$          & $87.06$    \\
					
					ITL-NET \cite{ltl-icml-2022} & $83.90$         & $78.90$      & $94.80$      & $80.10$          & $84.40$    
					& $87.10$         & $\underline{83.30}$      & $96.10$      & ${79.30}$          & $86.40$    \\
					
					\midrule

					VAUE (ours)  & $\textbf{85.82}$   & $79.37$ & $94.99$ & $\textbf{84.20}$ & $\textbf{86.10}$ 
					& ${88.30}$   & $\textbf{84.11}$ & $95.95$ & $\textbf{86.13}$ & $\textbf{88.62}$ \\
					\bottomrule[1.5pt]
				\end{tabular}
				\footnotesize
				The best result is in bold face. Underlined ones represent the second-best results. RSC* denotes the reproduced results from pAdaIN \cite{padin-cvpr-2021}.
	\end{threeparttable}}}}
\end{table}

% res18 pacs 70

\begin{table}[t]
	\renewcommand{\arraystretch}{1.}
	\centering
	{
		\caption{Ablation Study on PACS dataset.}
		\label{table:ablation}
		\setlength{\tabcolsep}{1.4mm}{\scalebox{0.75}{
				\begin{threeparttable}
					\begin{tabular}{l | c c c c |c | c c c c | c }
						\toprule[1.5pt]
						\multicolumn{1}{ l |}{Backbone} & \multicolumn{5}{c |}{ResNet-18} & \multicolumn{5}{c }{ResNet-50} \\
						\midrule
						Domain  &   A  & C   &   P   &  S    &   Avg.    &   A  & C   &   P   &  S    &   Avg. \\
						\midrule			
						VAUE & $\textbf{85.82}$   & $\textbf{79.37}$ & $\underline{94.99}$ & $\textbf{84.20}$ & $\textbf{86.10}$ 
						& $\textbf{88.30}$   & $\textbf{84.11}$ & $\underline{95.95}$ & $\textbf{86.13}$ & $\textbf{88.62}$ \\
						%					\multicolumn{11}{c} {\it{Ablation Study}} \\
						%					\midrule
						VAUE w/o VA& ${83.04}$   & $77.35$ & $\textbf{95.09}$ & ${79.85}$ & ${83.83}$ 
						& ${88.13}$   & $83.48$ & $\textbf{96.97}$ & ${82.40}$ & ${87.75}$ \\
						VAUE w/o EC& $\underline{85.77}$   & $\underline{78.77}$ & $94.51$ & ${82.88}$ & $\underline{85.48}$ 
						& ${88.12}$   & $\underline{83.70}$ & $95.03$ & ${84.44}$ & ${87.82 }$ \\
						VAUE w/o CD& ${85.27}$   & $78.03$ & ${94.89}$ & ${83.30}$ & ${85.37}$ 
						& $\underline{88.27}$   & $83.13$ & $95.14$ & $\underline{85.85}$ & $\underline{88.10}$ \\
						VAUE w/o UE& ${85.63}$   & $78.36$ & $94.13$ & $\underline{83.41}$ & ${85.38}$ 
						& ${87.23}$   & $82.42$ & $94.57$ & ${85.59}$ & ${87.45}$ \\
						\bottomrule[1.5pt]
					\end{tabular}
					\footnotesize
					The best result is in bold face. Underlined ones represent the second-best results. 
	\end{threeparttable}}}}
\end{table}

\section{Conclusion}

In this paper, we design the visual alignment module for dealing with the covariate shift by aligning the distribution of image style to a common Gaussian distribution. Another uncertainty-guided ensemble strategy is proposed to deal with the conditional shift between training domains and test samples by a dynamic adjustment. Experiment results show the excellent performance of the proposed VAUE.

%\newpage

\end{document}